\documentclass[twocolumn]{article}
\pdfoutput=1
\usepackage{amsmath}
\usepackage{amsfonts}
\usepackage{amssymb}
\usepackage[english]{babel}
\usepackage[export]{adjustbox}
\usepackage{bm} 
\usepackage{textcomp} 
\usepackage{gensymb} 
\usepackage{interval} 
\usepackage{xcolor} 
\usepackage{graphicx}
\usepackage{float}
\usepackage[caption=false]{subfig}
\usepackage{ragged2e}
\usepackage[acronym, toc]{glossaries}
\usepackage{acro}
\usepackage{capt-of}
\usepackage{makecell}
\usepackage[breaklinks=true,
            hidelinks,
            pdfauthor={Alex Halle},
            pdftitle={An AI-Assisted Design Method for Topology Optimization Without Pre-Optimized Training Data},
            pdfkeywords={deep learning, topology optimization, artificial neural networks, ai-assisted design}]{hyperref}
\usepackage{xurl}

\usepackage{csquotes}
\usepackage[
    backend=biber,
    style=numeric,
    natbib=true,
    sorting=none,
    url=false,
    doi=true,
    eprint=false
]{biblatex}
\usepackage{algorithm}
\usepackage{algpseudocode}
\usepackage{dblfloatfix}
\usepackage{authblk}
\usepackage{listings} 
\usepackage{multirow} 
\usepackage{booktabs} 
\usepackage{multirow, booktabs, tabularx} 
\usepackage{filemod} 

\newcolumntype{L}{>{\hspace{0pt}\RaggedRight}X}

\usepackage{xifthen}

\makeatletter
\renewcommand*\env@matrix[1][c]{\hskip -\arraycolsep
  \let\@ifnextchar\new@ifnextchar
  \array{*\c@MaxMatrixCols #1}}
\makeatother

\renewcommand{\vec}[1]{\boldsymbol{\mathbf{\MakeLowercase{#1}}}} 
\newcommand{\vecE}[1]{\MakeLowercase{#1}}           
\newcommand{\mat}[1]{\boldsymbol{\mathbf{\MakeUppercase{#1}}}}   
\newcommand{\matE}[1]{\MakeLowercase{#1}}           
\newcommand{\T}{\mathsf{T}}                         
\newcommand{\red}[1]{{#1_\mathrm{red}}}               

\newcommand{\matF}[2]{%
  \ifthenelse{\isempty{#2}}%
    {{\mat{#1}}}
    {{\mat{#1}_{\mathrm{#2}}}}
}
\newcommand{\vecF}[2]{%
  \ifthenelse{\isempty{#2}}%
    {{\vec{#1}}}
    {{\vec{#1}_{\mathrm{#2}}}}
}
\newcommand{\matEF}[2]{%
  \ifthenelse{\isempty{#2}}%
    {{\matE{#1}}}
    {{\matE{#1}_{\mathrm{#2}}}}
}
\newcommand{\vecEF}[2]{%
  \ifthenelse{\isempty{#2}}%
    {{\vecE{#1}}}
    {{\vecE{#1}_{\mathrm{#2}}}}
}

\newcommand{\trainingduration}{$T_\mathrm{P}~=~3~\mathrm{h}\  (2{:}56{:}32)$}  
\newcommand{\trainingsubdivision}{21~\%, 7~\%, 45~\%, 26~\%}  
\newcommand{\timeCONVperGeometry}{$1.9~\mathrm{s}$}           
\newcommand{\timePENperGeometry}{$8.4~\mathrm{ms}$}           
\newcommand{\timeBEP}{$5612$}                                 
\newcommand{\timeCONVslowerPEN}{$225$}                        
\newcommand{\exponentialsmoothingfactor}{$0.862$}             

\newcommand{\batchSizesperLevel}{$\begin{bmatrix}2048 \\ 512 \\ 128 \\ 8 \end{bmatrix}$} 

\newcommand{\msemean}{$0.093$}  
\newcommand{\mseSD}{$0.07$}     
\newcommand{\maemean}{$0.114$}  
\newcommand{\maeSD}{$0.075$}    
\newcommand{\kappaOOImean}{$79.7$~\%} 
\newcommand{\kappaOOISD}{$10.1$~\% }
\newcommand{\kappamean}{$86.3$~\%} 
\newcommand{\kappaSD}{$8.0$~\%} 

\newcommand{\dinp}{d_{\mathrm{inp}}}
\newcommand{\Xmin}{\vecE{X}_{\mathrm{min}}}
\newcommand{\Wp}{{\mat{W}_{\mathrm{p}}}}  
\newcommand{\Gp}{{\mat{G}_{\mathrm{p}}}}  

\newcommand{\EvaC}{F}

\newcommand{\ROned}{\mathcal{R}_{1d}}
\newcommand{\RTwod}{\mathcal{R}_{2d}}

\newcommand{\IR}{\mat{I}_{\mathrm{R}}}
\newcommand{\Mis}{{M_{\mathrm{is}}}}
\newcommand{\Mtar}{{M_{\mathrm{tar}}}}
\newcommand{\Rk}{{\vec{R}_{\mathrm{k}}}}
\newcommand{\Rkx}{\mat{R}_{\mathrm{k,x}}}
\newcommand{\Rky}{\mat{R}_{\mathrm{k,y}}}
\newcommand{\Rs}{{\vec{R}_{\mathrm{s}}}}
\newcommand{\RsF}{{\vecE{R}_{\mathrm{s,F}}}}
\newcommand{\Rsx}{\mat{R}_{\mathrm{s,x}}}
\newcommand{\Rsy}{\mat{R}_{\mathrm{s,y}}}

\newcommand*{\hl}[1]{\emph{#1}} 
\renewcommand{\and}{\\}

\algnewcommand\Or{\textbf{or}}

\intervalconfig{left open fence=(}
\intervalconfig{right open fence=)}

\providecommand{\keywords}[1]
{
  \small
  \textbf{\textit{Keywords---}} #1
}

\DeclareAcronym{AD}{
    short   = AD,
    long    = Automatic Differentiation
}
\DeclareAcronym{AI}{
    short   = AI,
    long    = Artificial Intelligence
}
\DeclareAcronym{ANN}{
    short   = ANN,
    long    = Artificial Neural Network,
}
\DeclareAcronym{BEP}{
    short   = BEP,
    long    = Break Even Point
}
\DeclareAcronym{DL}{
    short   = DL,
    long    = Deep Learning,
}
\DeclareAcronym{FEM}{
    short   = FEM,
    long    = Finite Element Method
}
\DeclareAcronym{GAN}{
    short   = GAN,
    long    = Generative Adversarial Network
}
\DeclareAcronym{mae}{
    short   = $mae$,
    long    = mean absolute error
}
\DeclareAcronym{mse}{
    short   = $mse$,
    long    = mean squared error
}
\DeclareAcronym{ML}{
    short   = ML,
    long    = Machine Learning
}
\DeclareAcronym{PEN}{
    short   = PEN,
    long    = Predictor-Evaluator-Network
}
\DeclareAcronym{ReLU}{
    short   = ReLU,
    long    = Rectified Linear Unit
}
\DeclareAcronym{ResNet}{
    short   = ResNet,
    long    = Residual Network
}
\DeclareAcronym{PReLU}{
    short   = PReLU,
    long    = Parametric Rectified Linear Unit
}
\DeclareAcronym{SD}{
    short   = SD,
    long    = Standard Deviation
}
\DeclareAcronym{SIMP}{
    short   = SIMP,
    long    = ``Solid Isotropic Material with Penalization''
}
\DeclareAcronym{SQP}{
    short   = SQP,
    long    = Sequential Quadratic Programming
}
\DeclareAcronym{TO}{
    short   = TO,
    long    = Topology Optimization,
}
\DeclareAcronym{top88}{
    short   = top88,
    long    = 88 lines of code
}

\begin{document}
    \renewcommand*{\Affilfont}{\normalsize\normalfont}

\title{An AI-Assisted Design Method for Topology Optimization Without Pre-Optimized Training Data}

\author[*,1, 2]{Alex Halle}
\author[1]{L. Flavio Campanile}
\author[1, 2]{Alexander Hasse}

\affil[1]{Professorship Machine Elements and Product Development, Chemnitz University of Technology, Germany}
\affil[2]{\textit {\{alex.halle, alexander.hasse\}@mb.tu-chemnitz.de}}

\maketitle

    \begin{abstract}
Topology optimization is widely used by engineers during the initial product development process to get a first possible geometry design. The state-of-the-art is the iterative calculation, which requires both time and computational power. Some newly developed methods use artificial intelligence to accelerate the topology optimization. These require conventionally pre-optimized data and therefore are dependent on the quality and number of available data.

This paper proposes an AI-assisted design method for topology optimization, which does not require pre-optimized data. The designs are provided by an artificial neural network, the predictor, on the basis of boundary conditions and degree of filling (the volume percentage filled by material) as input data. In the training phase, geometries generated on the basis of random input data are evaluated with respect to given criteria. The results of those evaluations flow into an objective function which is minimized by adapting the predictor's parameters.

After the training is completed, the presented AI-assisted design procedure supplies geometries which are similar to the ones generated by conventional topology optimizers, but requires a small fraction of the computational effort required by those algorithms. We anticipate our paper to be a starting point for AI-based methods that requires data, that is hard to compute or not available.
\end{abstract}

\keywords{deep learning, topology optimization, artificial neural networks, ai-assisted design}

    \section{Introduction}
\label{sec:Introduction}

In \ac{TO}, the material distribution over a given design domain is optimized by minimizing a certain objective function while fulfilling specified restrictions \citep{Sigmund2013}. In most cases, the optimization problem is solved in a mathematical way by means of a suitable search algorithm.

The present contribution deals with the solution of \ac{TO} problems by means of \ac{AI} techniques.
State-of-the-art research in this area require optimal structures on a data basis obtained by conventional \ac{TO}. For this reason, they are subject to several limitations which affect those techniques, such as large computational effort and problematic handling of multi-modal formulations. 
The approach proposed here aims at removing those drawbacks by generating all the artificial knowledge required for the optimization during the learning phase, with no need of relying on pre-optimized results.

        \subsection{Topology Optimization}
\label{sec:Topology Optimization}

In this work, only the case of mono-material topology optimization will be considered. The material of which the structure is to be build is a constant of the problem and geometry remains as unknown.

In the case of stiffness optimization, a scalar measure of structural compliance is typically chosen as the function to be minimized. 
In addition, the condition that a given quantity of material is used over the design domain must be fulfilled. 
This material quantity is expressed as fraction of the maximum possible amount of material (degree of filling). 
Minimization of compliance results in maximizing the stiffness. The available design domain, the static and kinematic boundary conditions for the regarded load cases as well as strength thresholds are typically considered as restrictions.

This paper will also focus on stiffness optimization, although the presented method is of general validity and could be applied to optimization with different objective functions or restrictions.

There are numerous possible approaches to \ac{TO} \citep{Sigmund2013}. In the \ac{SIMP} approach according to Bends{\o}e \citep{Bendsoe2003} the design domain is divided into elements. For each of those elements the contribution to the overall stiffness of the structure is scaled with a factor to be determined.

The \ac{SIMP} approach is able to provide optimized geometries for many practical cases by means of an iterative process. Each iteration involves computationally intensive operations: the most critical ones are assembling  the stiffness matrix and solving the system's equation. When restrictions are involved, such as stress restrictions, the complexity of the optimization problem increases \citep{Picelli2018,Lee2012}.

        \subsection{Artificial Neural Networks}
\label{sec:Artificial_Neural_Networks}

\Acp{ANN} belong to the area of \ac{ML}, which, in turn, are assigned to \ac{AI}. 
\Acp{ANN} are able to learn and execute complex procedures, which has led to remarkable results in recent years. For example, \acp{ANN} are able to recognize the objects shown on pictures by their shape and color or beat world champions in the board game ``GO'' \citep{Redmon2015,Deepmind2019}.

The development of \acp{ANN} is progressing steadily, on the one hand due to the continuously better available computing power and on the other hand due to the discovery of new possibilities to improve the learning process.

\begin{figure}[ht]
    \centering
    \includegraphics[scale=1]{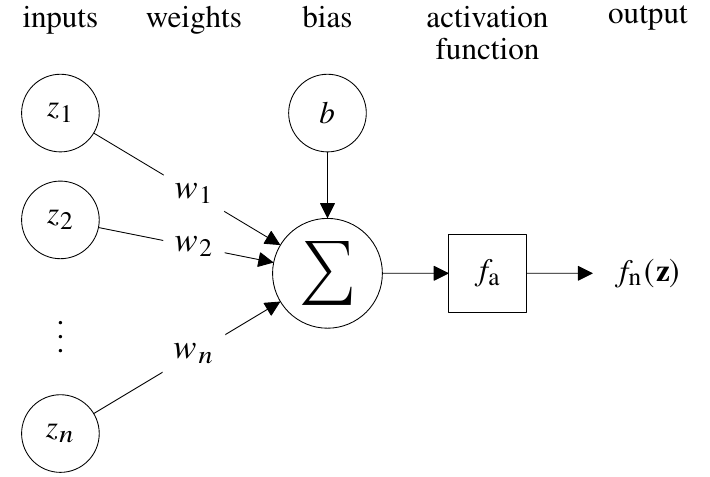}
    \caption{Graphical representation of a single neuron}
    \label{fig:Graphical representation of a single neuron}
\end{figure}

\Acp{ANN} or, more precisely, feedforward neural networks consist of layers connected in sequence. 
These layers contain so-called neurons \citep{Karayiannis1993}. A neuron (see Figure \ref{fig:Graphical representation of a single neuron}) is the basic element of an \ac{ANN}. The combination of all layers is also called a network.

The neuron receives $n$ inputs (here given as vector $\vec{Z}$), which are linearly combined and added to a bias value $b$ and passed as argument to an activation function $f_a$
\begin{equation}
    \centering
    f_\mathrm{n}(\vec{Z}) = f_\mathrm{a}(\vec{W}^\T \vec{Z}+b)
    .
    \label{eq:1}
\end{equation}

The coefficients of the linear combination, collected in the vector $\vec{W}$, are called weights.

It is usual that several neurons have the same input. All neurons with the same inputs are grouped together in one layer (also called \hl{fully connected layer}). Since one single output is supplied by each neuron, each layer with $m$ neurons also produces $m$ outputs and the weights become matrices $\mat{W} \in \mathbb{R}^{n \times m}$. The outputs of a layer (except the last layer) serve as inputs for the following layer. The first layer is called input layer $f^{(1)}$ and the last layer is called output layer $f^{(n_\mathrm{L})}$. Any layer whose input and output values are not accessible to the user is called hidden layer $f^{(2, \ldots, n_\mathrm{L}-1)}$. Each layer, for example the layer $f^{(2)}$, has its own weights $\mat{W}^{(2)}$ and biases $\vec{b}^{(2)}$.

The number of layers $n_\mathrm{L}$ is also named depth of the network, which also originate the attribute ``deep'' in the term \ac{DL}. The term \ac{DL} is generally used for \acp{ANN} with several hidden layers. The presence of several layers makes it possible to map a more complex transfer behavior between the input and the output layer.

The functional relationship
\begin{equation}
    \begin{aligned}
        y &= f_\mathrm{ANN}(\vec{Z},\mat{W}^{(1, \ldots, n_\mathrm{L})},\vec{b}^{(1, \ldots, n_\mathrm{L})}) \\
          &= f^{(n_l-1)}
            \left(
                f^{(\ldots)}
                \left(
                    f^{(2)}
                    \left(
                        f^{(1)}
                        \left(
                            \vec{Z} 
                        \right) 
                    \right) 
                \right) 
            \right)
    \end{aligned}
    \label{eq:2}
\end{equation}
realized by the \ac{ANN} depends on the weights $\mat{W}$ and on the biases $\vec{b}$, which are adjusted in the context of so-called \hl{training} or \hl{learning} according to certain algorithms (learning algorithms). The learning algorithm used consists in the gradient-based minimization of a scalar value termed \hl{error} or \hl{loss}, which is obtained from the deviations of the actual outputs from given target outputs. The values which describe the network's architecture and do not undergo any change during training, like the number of neurons in a layer, are termed \hl{hyperparameters}.

In addition to the fully connected layers there are \hl{convolutional layers}. These layers use convolution in place of the linear combination \eqref{eq:1}. Here the trainable weights, also called convolution kernel, are convoluted with the input of the layer and produce an output which is passed to the next layer. This process is efficient for grid-like data structures and is therefore used for many modern image applications \citep{Goodfellow2016}.

The output of an \ac{ANN} is henceforth referred to as \hl{prediction}. Further details on the learning of an \ac{ANN} can be found in the specific literature, for example \citep{Goodfellow2016,Basheer2000}.

        \subsection{\Ac{ANN}-based \acl{TO}}
\label{sec:DL_Based_Topology_Optimization}

\Ac{DL}-based \ac{TO}, by predicting the geometry through \acp{ANN}, aims to deliver optimized results in only a fraction of the time required by conventional optimization, by moving the computationally intensive part to the training algorithm, which is executed only once. The results provided by the trained \acp{ANN} can then be used directly, refined with conventional methods or adapted to the desired structure size.

There are already some attempts in this area. The majority used many thousands topology-optimized geometries as training datasets for the \ac{ANN} \citep{Yu2019, Rawat2019, Zhang2019, Sasaki2019, Malviya2020}.

In the case of \citep{Zhang2019} 80,000 optimized datasets based on the \ac{top88} \citep{Andreassen2011} were used for the training of the neural network.

Banga et al. used an approach in which intermediate results of conventional \ac{TO} are the basis for the training datasets for the \ac{ANN} \citep{Banga2018}.

Nie et al. used a \ac{GAN} \cite{Goodfellow2014} and some physical fields over the initial domain, like strain energy density or von Mises Stress, to predict the geometries. The model was trained with 49.078 conventionally topology-optimized datasets \cite{Nie2020}. 

Yamasaki et al. \cite{Yamasaki2021} and Cang et al. \cite{Cang2018} achieved great results using data-driven approaches. In the case of Cang two cases were trained and tested. In the first case the direction of a single load could be is changed. The second case had variable directions as well as the positions of the load. 
Yamasaki's \ac{ANN} needed to be trained for each new boundary condition. Only the volume fraction is variable. 

A different approach was presented by Chandrasekhar and Suresh, where the \ac{ANN} does not generate the whole geometry but only a density value at given x and y coordinates \cite{Chandrasekhar2021}. 

Although such \ac{ANN} topology optimization procedures are able to perform the above-mentioned task of a fast and direct generation of optimized geometries, the predictions undergo some restrictions.

Since topology-optimized training data are used, and the generation of these data with conventional methods is very time-consuming, the number of training data sets which can be considered is limited. In the case of \citep{Yu2019} 100,000 data sets were generated and this took about 200h. Another 8h were needed for training.
This limitation affects the accuracy in the prediction of unknown geometries (i.e. geometries which were not used within the training) negatively. For example in the paper \citep{Yu2019} about 3.4~\% of the generated geometries are not connected (with theoretically infinite compliance) and are therefore not usable.

This paper investigates the possibility, which differs from the state of the art, to train an \ac{ANN} without the use of topology-optimized data sets. The generation of training data sets and the training itself are merged in one single procedural step.

This makes it possible to process a much larger amount of data sets for the training in a much shorter time.
And since the compliance is calculated during the training the \ac{ANN} learns to avoid undesirable results.

The state-of-the-art procedures require the use of large number of optimized data sets. These data sets must be optimal to be suitable as training data. Depending on the optimization formulation, this may be not the case, as local minima and convergence problems may occur. A method which doesn't use optimized data sets is not subject to these restrictions.

    \section{Method}
\label{sec:Method}

The presented method is based on an \ac{ANN} architecture called \ac{PEN}, which was developed by the authors for this purpose. The predictor is the trainable part of the \ac{PEN} and its task is to generate---based on input data sets---optimized geometries.

As mentioned, unlike the state-of-the-art methods mentioned above, no pre-optimized topology-optimized data sets are used in the training. The geometries used for the training are created by the predictor itself on the basis of randomly generated input data sets and evaluated by the remaining components of the \ac{PEN}, called evaluators.

The evaluators perform mathematical operations. Other than the predictor, the operations performed by the evaluators are pre-defined and do not change during the training. 

Each evaluator assesses the outputs of the predictor with respect to a certain criterion and returns a corresponding scalar value as measure of the criterion's fulfillment. This fulfillment is the \hl{loss} or the \hl{error} of this evaluator. 
A scalar function of the evaluator outputs (objective function $J$, see section \ref{sec:Objective function}) combines the individual losses.

During the training the objective function computed for a set of geometries (batch) is minimized by changing the predictor's trainable parameters, see section \ref{sec:Predictor}. In this way, the predictor learns how to produce optimized geometries.

The predictor, the individual evaluators, their tasks and their way of operation are explained in detail in the following sections.

        \subsection{Basic Definitions}
\label{sec:Basic_Definitions}

\begin{figure}[!ht]
    \centering
    \includegraphics[scale=1]{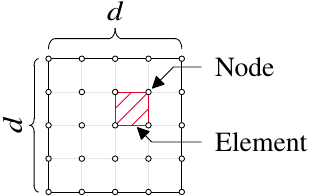}
    \caption{Element and Nodes}
    \label{fig:Element and Nodes}
\end{figure}

In topology optimization, the design domain is typically subdivided in elements by appropriate meshing. In Figure \ref{fig:Element and Nodes}, elements (with one element hatched) and nodes are visualized.

In this work we examined only square meshes with equal number of rows and columns. Although this method can be used for non-square and three-dimensional geometries.

The total number of elements in the 2d-case is
\begin{equation}
    d = d_x d_y,
\end{equation}
where $d_y$ is the number of rows and $d_x$ the number of columns (see Figure \ref{fig:Element and Nodes}). In the square case the number of rows and columns are equal $d_x = d_y = d$.

The $d^2$ design variables $\vecE{X}_i \ \{i=1, \ldots, d^2\}$, termed density values, scale the contributions of the single elements to the stiffness matrix. The density has the value one when the stiffness contribution of the element is fully preserved and zero when it disappears.

The density values are collected in a vector $\vec{X}$. In general the density values $\vecE{X}_i$ are defined in the interval $\interval{0}{1}$. In order to prevent possible singularities of the stiffness matrix, a lower limit value $\Xmin$ for the entries of $\vec{X}$ is set \citep{Bendsoe2003}:
\begin{equation}
    0 < \Xmin \leq \vecE{X}_i \leq 1 \text{,} \quad i = 1,2, \ldots, d^2
    .
    \label{eq:9}
\end{equation}

The vector of design variables $\vec{X}$ can be transformed to a square matrix $\mat{X}_\mathrm{M}$ of order $d$ by using the $\RTwod$ operator:

\begin{equation}
    \begin{array}{l}
        \RTwod (\vec{X}) = \mat{X}_\mathrm{M} = ({\matE{X}_\mathrm{M}}_{ij}) \\
        \text{where} \ \matE{X}_{\mathrm{M}_{ij}} = \vecE{X}_{i+d(j-1)} \text{,} \\
        i=1,2, \ldots, d, \quad j=1,2, \ldots, d.
        \label{eq:R_2d}
    \end{array}
\end{equation}

Although a binary selection of the density is desired (discrete \ac{TO}, material present/not present), values between zero and one are permitted for algorithmic reasons (continuous \ac{TO}). To get closer to the desired binary selection of densities the so-called penalization can be used in the calculation of the compliance. The penalization is realized by an element-wise exponentiation of the densities by the penalization exponent $p > 1$  \citep{Sigmund2001}.

The arithmetic mean of all $\vecE{X}_{i}$ defines the degree of filling of the geometry
\begin{equation}
    \Mis = \frac{1}{d^2} \sum_{i=1}^{d^2} \vecE{X}_{i}
    .
    \label{eq:M_is}
\end{equation}
The target value $\Mtar$ is the degree of filling that is to be achieved by the predictor.

\begin{figure}[!ht]
    \centering
    \includegraphics[scale=1]{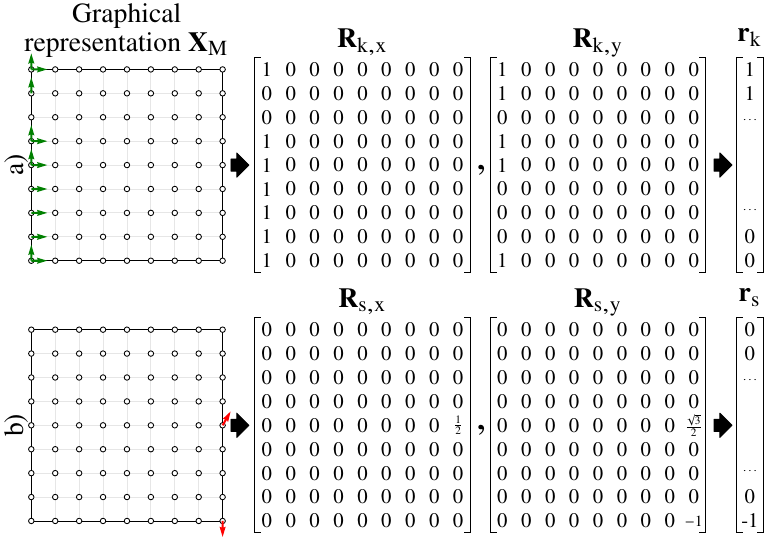}
    \caption{Matrix representation of a) kinematic boundary conditions and b) static boundary conditions}
    \label{fig:Matrix representation of the boundary conditions}
\end{figure}

The kinematic boundary conditions are stored in two $(\dinp+1)\times(\dinp+1)$ boolean matrices  $\Rkx$ and $\Rky$. In Figure \ref{fig:Matrix representation of the boundary conditions}, which shows an overview of how the boundary condition are handled, as well as in following figures, the green arrows represent the kinematic boundary conditions and the red ones the static boundary conditions. The entries of $\Rkx$ are set to one if the x-component of the displacement in the corresponding node is fixed, and zero otherwise. Analogously, the entries of $\Rky$ are set according to the fixed y-components of the displacements. Both matrices can be transformed into vectors with the $\ROned$ operator, which is the inverse of operator $\RTwod$, and then arranged in sequence so that the vector $\Rk \in \mathbb{R}^{\left({2 \dinp + 1}\right) ^2} $ is created.

Analogously to the kinematic boundary conditions, two $(\dinp+1)\times(\dinp+1)$ matrices $\Rsx$ and $\Rsy$ are firstly built on the basis of the static boundary conditions (visualized by red arrows). The x- and the y-components of the applied forces are placed, respectively, into the matrices in correspondence to their magnitude, while the remaining entries are set to zero. The matrices $\Rsx$ and $\Rsy$ are then converted into the vector $\Rs$.

Investigations showed that the training speed could be increased, for high-resolution geometries, by dividing the training into levels with increasing resolution. Since smaller geometries are trained several orders of magnitude faster and the knowledge gained is also used for higher resolution geometries, the overall training time is reduces compared to the training that uses only high-resolution geometries. The levels are labeled with the integer number $\Lambda$.

Increasing $\Lambda$ by 1 results in doubling the number $d$ of row or columns of the design domain's mesh. This is done by quartering the elements of the previous level. In this way, the nodes of the previous level are kept in the new level. The number of row or columns at the first level is denoted as $\dinp$.

\begin{figure}[!ht]
    \centering
    \includegraphics[scale=1]{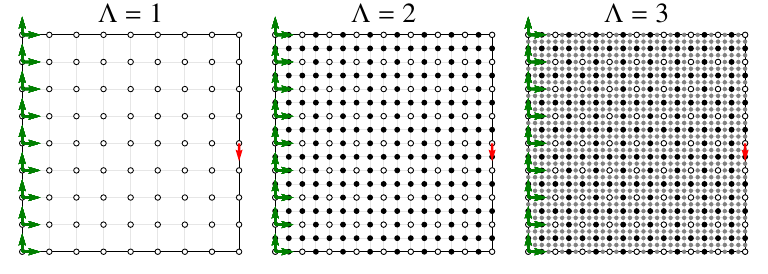}
    \caption{Nodes and elements for different levels $\Lambda$. The green arrows represent the kinematic boundary conditions. The red arrows represent the static boundary condition.}
    \label{fig:Nodes for different levels}
\end{figure}

The input data of the predictor includes the kinematic $\Rk$ and static $\Rs$  boundary conditions as well as the target degree of filling $\Mtar$. The output of the predictor is a geometry $\vec{X}$. Input data can be only defined at the initial level and do not change while the level is changed. Hence, new nodes cannot be subject to static or kinematic boundary conditions (see Figure \ref{fig:Nodes for different levels}). The change of level occurs after a certain condition, which will be described later, is fulfilled.

        \subsection{Predictor}
\label{sec:Predictor}

\begin{figure}[htb]
    \centering
    \includegraphics[width=\linewidth]{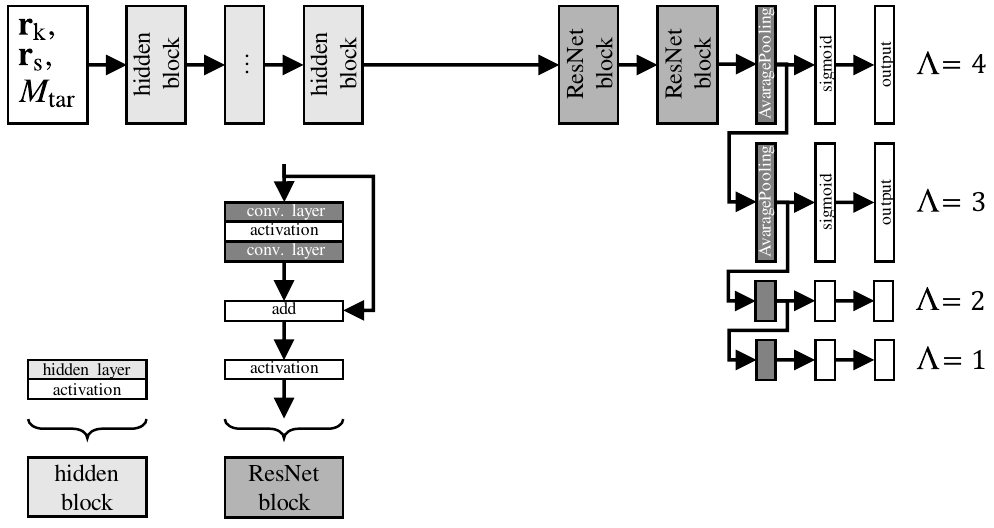}
    \caption{Predictor's topology simplified}
    \label{fig:predictor_topology_simplified}
\end{figure}

The predictor is in charge of creating an optimized geometry for a given input data set. Its \ac{ANN}-architecture consist of multiple hidden layers, convolutional layers and an output layer with $d^2$ neurons (see Figure \ref{fig:predictor_topology_simplified}).

As activation function $f_\mathrm{a}(z)$ in the hidden and convolutional layers the \hl{\ac{PReLU}} function is being used \citep{Abadi2015}. The \hl{PReLU} function is the equivalent of the \hl{\ac{ReLU}} function \citep{Abadi2015}
\begin{equation}
    f_{\mathrm{ReLU}}(z) = max(0, z) =
            \begin{cases}
                0 & \text{if $z < 0$,} \\
                z & \text{otherwise,}
            \end{cases}
    \label{eq:ReLU}
\end{equation}
with the difference of a variable negative slope $\alpha$, which can be adapted during training:
\begin{equation}
    f_\mathrm{a}(z) = f_{\mathrm{PReLU}}(z) =
            \begin{cases}
                \alpha z & \text{if $z < 0$,} \\
                z & \text{otherwise.}
            \end{cases}
    \label{eq:PReLU}
\end{equation}

The sigmoid function
\begin{equation}
    \mathrm{sig}(z) = \frac{1}{1 + e^{-z}}
    \label{eq:8}
\end{equation}
is well suited as activation function for the output layer because it provides results in the interval $\interval[open]{0}{1}$, see Figure \ref{fig:Sigmoid, ReLU and PReLU function}. This makes the predictor's output directly suitable to describe the density values of the geometry.
\begin{figure}[htb]
    \centering
    \includegraphics[scale=1]{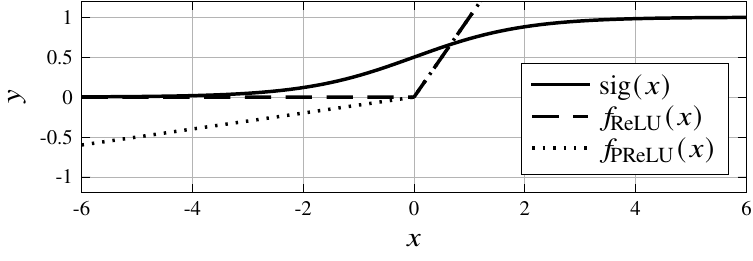}
    \caption{Sigmoid, ReLU and PReLU function}
    \label{fig:Sigmoid, ReLU and PReLU function}
\end{figure}

All parameters that can be changed during training, like the bias, the slope of the \ac{PReLU} as well as the weights of the hidden layers will be generally referred to as \hl{trainable parameters} in the following. They are collected in the matrix $\Wp$. The operations performed by the predictor can be represented by a function $f_\mathrm{p}$:
\begin{equation}
    \vec{X} = f_\mathrm{p}(\Wp, \Rk, \Rs, \Mtar)
    .
    \label{eq:10}
\end{equation}

The predictor's topology is shown in Figure \ref{fig:predictor_topology_simplified} in a simplified form.

Here the data flow through the predictor as well as output layers for different level $\Lambda$ can be seen. An input data set (top left) is processed by several successive hidden layers and then passed on to some \ac{ResNet}-blocks.
In order to reduce the resolution to a lower $\Lambda$, \hl{average pooling} is used.

In Figure \ref{fig:predictor_topology_simplified} the \hl{hidden block} is the combination of a hidden or fully connected layer and an activation function call. The \ac{ResNet}-block is the combination of two (convolutional) layers and a shortcut that is added as a bypass to the output of the layers. The \ac{ResNet}-block allows for faster learning but also reduces the error \cite{He2015}.

For subsequent levels, the outputs of the last convolutional block of the previous layer and the outputs of the last hidden block of the first level are added together and then, after an additional convolutional block, converted to the desired output dimension.

        \subsection{Evaluator: Compliance}
\label{sec:Evaluator: Compliance}

The task of the compliance evaluator is the computation of the global mean compliance. For this purpose, an algorithm based on \ac{FEM} \citep{Sigmund2001} is used. 
The global mean compliance
\begin{equation}
    c = f_{\mathrm{eva,c}}(\vec{X}, \Rk, \Rs)
    \label{function_eva_c}
\end{equation}
is defined according to \citep{Sigmund2001} as
\begin{equation}
    c = \vec{U}^\T \mat{K} \vec{U} = \vec{U}^\T \vec{F}
    \label{eq:c=U^T K F=U^T F}
\end{equation}
with $\mat{K}$ as the stiffness matrix,  $\vec{F}$ as the force vector and  $\vec{U}$ as the displacement vector. The compliance has the dimension of energy. As usual in literature \citep{Sigmund2001, Andreassen2011} in the following the units will be omitted for the sake of simplicity.

As already explained, the static boundary conditions vector $\Rs$ consists first of x-entries and then y-entries. Since the degrees of freedom of the stiffness matrix are arranged in alternate way (one x-entry and one y-entry), the force vector is to be built accordingly. In order to transform the static boundary condition vector $\Rs$ into the force vector $\vec{F}$, the number of nodes
\begin{equation}
    l = (d+1)^2
\end{equation}
and a collocation matrix $\IR$
\begin{equation}
    \IR =
    \begin{cases}
        \matE{I}_{\mathrm{R}_{2i,i}} = 1 & \text{where }  i=\{0, \ldots ,l - 1 \}, \\
        \matE{I}_{\mathrm{R}_{2(j - l)+1,j}} = 1 & \text{where } j=\{l, \ldots ,2l - 1 \}, \\
        \matE{I}_{\mathrm{R}} = 0 & \text{otherwise}
    \end{cases}
    \label{eq:I_R)}
\end{equation}
are required.
The force vector is then obtained as follows:
\begin{equation}
    \vec{F} = \IR \Rs
    .
    \label{eq:F_(R_S)}
\end{equation}
The system's equations write
\begin{equation}
    \vec{F} = \mat{K} \vec{U}
    \label{eq:F=KU}.
\end{equation}
The stiffness matrix $\mat{K}$ depends linearly on the geometry $\vec{X}$ and is expressed by
\begin{equation}
    \mat{K} = \sum_{i = 1}^{d^2} \vecE{X}_i^p \mat{K}_i
    \label{eq:K_calculation}
\end{equation}
where the matrices $\mat{K}_i$ are the unscaled contributions of the single elements to the stiffness matrix. The penalization exponent $p$ achieves the desired focusing of the geometry towards the limits values $\Xmin$ and $1$ as described in section \ref{sec:Basic_Definitions}.

The stiffness matrix $\mat{K}$ is then reduced by removing the columns and rows corresponding to the fixed degrees of freedom according to the kinematic boundary conditions. The result is the reduced stiffness matrix $\red{\mat{K}}$, which then can be inverted. The reduced force vector $\red{\vec{F}}$ is determined according to the same principle. From the reduced equation
\begin{equation}
    \red{\vec{F}} = \red{\mat{K}} \red{\vec{U}}
    \label{eq:F_red=K_red U_red}
\end{equation}
the reduced displacement vector is obtained as
\begin{equation}
    \red{\vec{U}} = \red{\mat{K}}^{-1} \red{\vec{F}}.
\end{equation}

The reduced global mean compliance $\red{c}$ is finally computed as follows:
\begin{equation}
    \red{c} = c = \red{\vec{U}}^\T \red{\mat{K}} \red{\vec{U}}.
    \label{eq:c=U_red^T * K_red * U_red}
\end{equation}

The calculation of the mean global compliance $c$ according to \eqref{eq:c=U^T K F=U^T F} or $\red{c}$ according to \eqref{eq:c=U_red^T * K_red * U_red} leads to the same result, since $\vec{U}$ at the fixed degrees of freedom vanishes and therefore have no effect on $c$.

        \subsection{Evaluator: Degree of filling}
\label{sec:Evaluator: Degree of filling}

The task of this evaluator is to determine the deviation of the degree of filling $\Mis$, see \eqref{eq:M_is}, from the target value $\Mtar$
\begin{equation}
    M = f_{\mathrm{eva,M}}(\vec{X}, \Mtar) = \left\lvert \Mtar - \Mis \right\rvert
    .
    \label{eq:M}
\end{equation}
By considering the filling degree's deviation $M$ in the objective function, the predictor is penalized proportionally to the extent of the deviation from the target degree of filling $\Mtar$.

        \subsection{Evaluator: Filter}
\label{sec:Evaluator: Filter}

The filter evaluator searches for checkerboard patterns in the geometry and outputs a scalar value $\EvaC \in \interval{0}{1}$ that points to the amount and extent of checkerboard patterns detected. These checkerboard patterns consist of alternating high and low density values of the geometry. They are undesirable because they do not reflect the optimal material distribution and are difficult to transfer to real parts. These checkerboard patterns exist due to bad numerical modelling \citep{Diaz1995}.

\begin{figure}[!ht]
    \centering
    \includegraphics[scale=1]{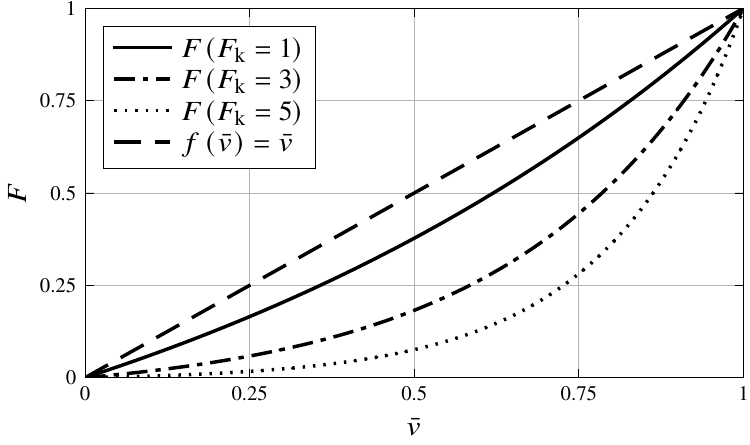}
    \caption{Influence of factor $\EvaC_{\mathrm{k}}$ on filter calculation}
    \label{fig:Influence of factor F_k on filter calculation}
\end{figure}

Several solutions for the checkerboard problem were developed in the framework of conventional topology optimization \citep{Sigmund1998}. In this work, a new strategy was chosen, which allows for inclusion of the checkerboard filter into the quality function. In the present approach, checkerboard patterns are admitted, but detected and penalized accordingly.
Since the type of implementation is fundamentally different, it is not possible to compare the conventional filter method with the filter evaluator.
With the matrix
\begin{equation}
    \mat{H} = \frac{1}{4}
    \begin{bmatrix}[r]
        1  & -1 & 1\\
        -1 & 0 & -1\\
        1  & -1 & 1
    \end{bmatrix}
    = (\matE{H}_{ij}) \in \mathbb{R}^{3 \times 3}
\end{equation}
a two-dimensional convolution operation (discrete convolution)
\begin{equation}
    \mat{V} (\vec{X}) =
    \left\lvert \RTwod (\vec{X}) \ast \mat{H} \right\rvert  =
    \left\lvert \mat{X}_\mathrm{M} \ast \mat{H} \right\rvert =
    (\matE{V}_{ij})
\end{equation}
is performed. In detail, the convolution operation is carried out as follows:
\begin{equation}
    \begin{array}{l}
        \matE{V}_{pq} =
        \sum_{i=1}^{3} \sum_{j=1}^{3}
        \left\lvert \matE{X}_{\mathrm{M}_{(p+i-1)(q+j-1)}} \matE{H}_{ij} \right\rvert  \\
        \text{with} \ p,q \in \{ 1, 2, \ldots , d-2 \}.
    \end{array}
\end{equation}
The convolution matrix $\mat{V}$ is visualized in Figure \ref{fig:Left - sample geometry, right - convolutional matrix V} for an exemplary case.
\begin{figure}[htb]
    \centering
    \includegraphics[width=\linewidth]{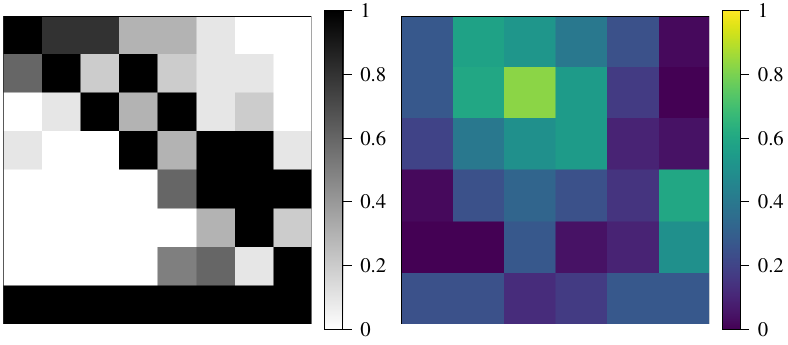}
    \caption{Left \textemdash{} sample geometry, right \textemdash{} convolutional matrix $\mat{V}$}
    \label{fig:Left - sample geometry, right - convolutional matrix V}
\end{figure}
The matrix $\mat{V}$ has high values in areas where the geometry has checkerboard patterns. A first indicator can be computed as mean value of the convolution matrix:
\begin{equation}
    \bar{\matE{V}} = \frac{1}{(d-2)^2} \sum_{i=1}^{d-2} \sum_{j=1}^{d-2} \matE{V}_{ij}
    .
\end{equation}
This indicator would already be sufficient to exclude geometries with checkerboard patterns but also penalizes good geometries without recognizable checkerboard patterns. Therefore, an improved indicator is formed on the basis of the mean value and with the help of the $e$-function, which is less sensitive to small mean values but nevertheless results in a corresponding penalization for large checkerboard patterns:
\begin{equation}
    \EvaC = f_{\mathrm{eva,\EvaC}} (\vec{X}) = \frac{e^{\bar{\matE{V}} \cdot \EvaC_{\mathrm{k}}} - 1}{e^{\EvaC_{\mathrm{k}}} - 1}
    .
\end{equation}
The parameter $\EvaC_{\mathrm{k}}$ controls the shape of the $\EvaC$-function (see Figure \ref{fig:Influence of factor F_k on filter calculation}).

        \subsection{Evaluator: Uncertainty}
\label{sec:Evaluator: Uncertainty}

When calculating the density values of the geometry $\vec{X}$, the predictor should, as far as possible, focus on the limit values $\Xmin$ and $1$ and penalize intermediate values. The deviation from this goal is expressed by the uncertainty evaluator with the scalar variable $P$ (uncertainty). This value increases if the predicted geometry deviates significantly from the limit values and thus penalizes the predictor. The uncertainty evaluator uses the normal distribution function
\begin{equation}
    f_g(x) = \frac{1}{\sqrt{2 \pi \sigma^2}} e^{-\frac{(x-\mu)^2}{2\sigma^2}}
    \label{eq:f_g}
\end{equation}
with $\sigma^2$ as the variance and $\mu$ as the expected value. The expected value is set to $\frac{1}{2}$, at which $P$ should have the maximum. In order for $P$ to be normalized (with $x=\frac{1}{2}$ the function should have the value $1$), the normal distribution function $f_g(x)$ is multiplied by the term $\sqrt{2 \pi \sigma^2}$.
\begin{equation}
    f_{g,n}(x) = \frac{1}{\sqrt{2 \pi \sigma^2}} e^{-\frac{(x-\frac{1}{2})^2}{2\sigma^2}}
    \cdot \sqrt{2 \pi \sigma^2} = e^{-\frac{(x-\frac{1}{2})^2}{2\sigma^2}}
    \label{eq:f_g,n}
\end{equation}
The resulting function $f_{g,n}$ is evaluated for all elements of the geometry. The mean value of the results provides the uncertainty:
\begin{equation}
    P = f_{\mathrm{eva,P}} (\vec{X}) = \frac{1}{d^2} \sum_{i=1}^{d^2} f_{g,n}(\vecE{X}_i)
    \label{eq:P}
\end{equation}
The variance $\sigma^2$ determines the width of the distribution function.

        \subsection{Quality function and objective function}
\label{sec:Objective function}

The task of the quality function is to combine all evaluator losses into one scalar. The following additional requirements must be considered:
\begin{itemize}
    \item The function should have a simple mathematical form, in order not to complicate the minimum search.
    \item The function must be monotonically increasing with respect to the evaluators' losses 
    \item The function contains coefficients to control the relative influence of the evaluators losses 
\end{itemize}
The most obvious variant fulfilling these criteria a linear combination of the losses. 
The problem with this choice consists in the different and variable order of magnitude of the compliance loss with respect to the other losses. 
For a given choice of the coefficients the relative influence of the losses changes for different parametrization and input data sets. 
To avoid this drawback, a quality function in the following form was chosen
\begin{equation}
    f_\mathrm{Q}(c, M, \EvaC) = (\alpha c + 1) \cdot (\beta M + 1) \cdot (\gamma \EvaC + 1) \cdot (\delta P + 1)
    \label{eq:f_Q}.
\end{equation}
The addition of the constant value prevent the quality function from being dominated by one loss when its value is close to zero.

For every single dataset one value of $f_\mathrm{Q}$ exists. Optimization on the basis of single datasets would require large computational effort and lead to instabilities of the training process (large jumps of the objective function output).
Therefore, a given number $b_n$ of datasets (batch) is used and the corresponding quality function values are combined in one scalar value, which works as objective function for the optimization that rules the training. The value of the objective function
\begin{equation}
    \begin{aligned}
            J &= \frac{1}{b_n} \sum_{i=1}^{b_n} f_{\mathrm{Q}_{i}} \\
            &= \frac{1}{b_n} \sum_{i=1}^{b_n} (\alpha c_i + 1) \cdot (\beta M_i +1) \cdot (\gamma \EvaC_i + 1) \cdot (\delta P_i + 1)
    \label{eq:J}
    \end{aligned}
\end{equation}
is calculated as the arithmetic mean of the quality function values obtained for the single datasets of the batch. Investigations showed that averaging the quality function outputs over numerous data sets stabilizes the training procedure. The disadvantage of this averaging is the possibility of forming prejudices. E.g. if one element is frequently present, then its frequency is also learned, even if the element's contribution to stiffness is in some cases small or non-existent.

        \subsection{Training}
\label{sec:Training}

\begin{figure}[ht]
    \centering
    \includegraphics[scale=1]{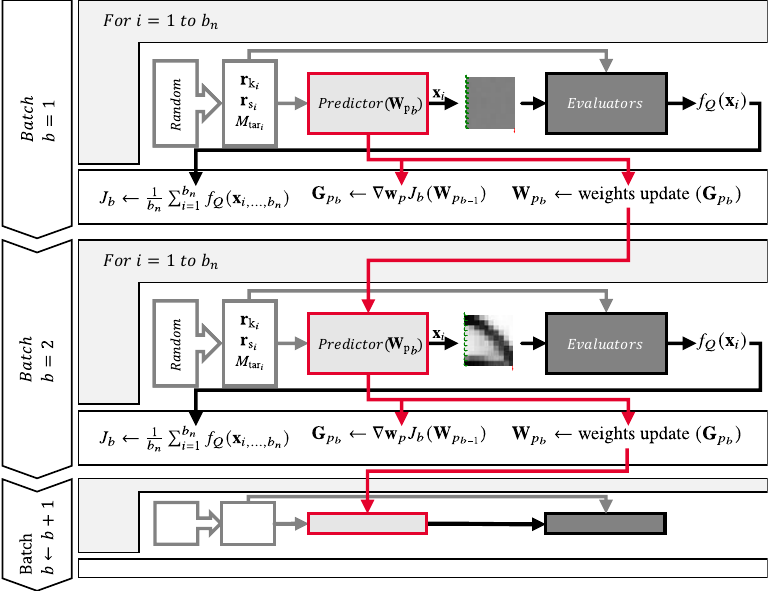}
    \caption{Structure of the \acf{PEN}}
    \label{fig:Method}
\end{figure}

The overview in Figure \ref{fig:Method} describes the training process for a single level. Here, it is visible that during a batch iteration the input data sets are calculated randomly and then passed to the predictor as well as to the evaluators.
Within one batch the input datasets are randomly generated and the predictor creates the corresponding geometries $\vec{X}_i$. Afterwards, the quality function is computed from the evaluators losses according to \eqref{eq:f_Q}. The objective function $J$ is then calculated for the whole batch. Then the gradient $\Gp_b$ of the objective function with respect to the trainable parameters is calculated. The trainable parameters of the predictor for the next batch are then adjusted according to the steepest-descent criterion to decrease the value of the objective function.

When the level increases, the predictor outputs a geometry with higher resolution and the process starts again at batch $b=1$.

It is important to stress that, unlike conventional topology optimization, the PEN method does not optimize the density values of the geometry, but only the weights of the predictor.

    \section{Application}
\label{sec:Application}

        \subsection{Implementation}
\label{sec:Implementation}

\begin{figure*}[tb]
    \centering
    \scalebox{1}{
        \includegraphics{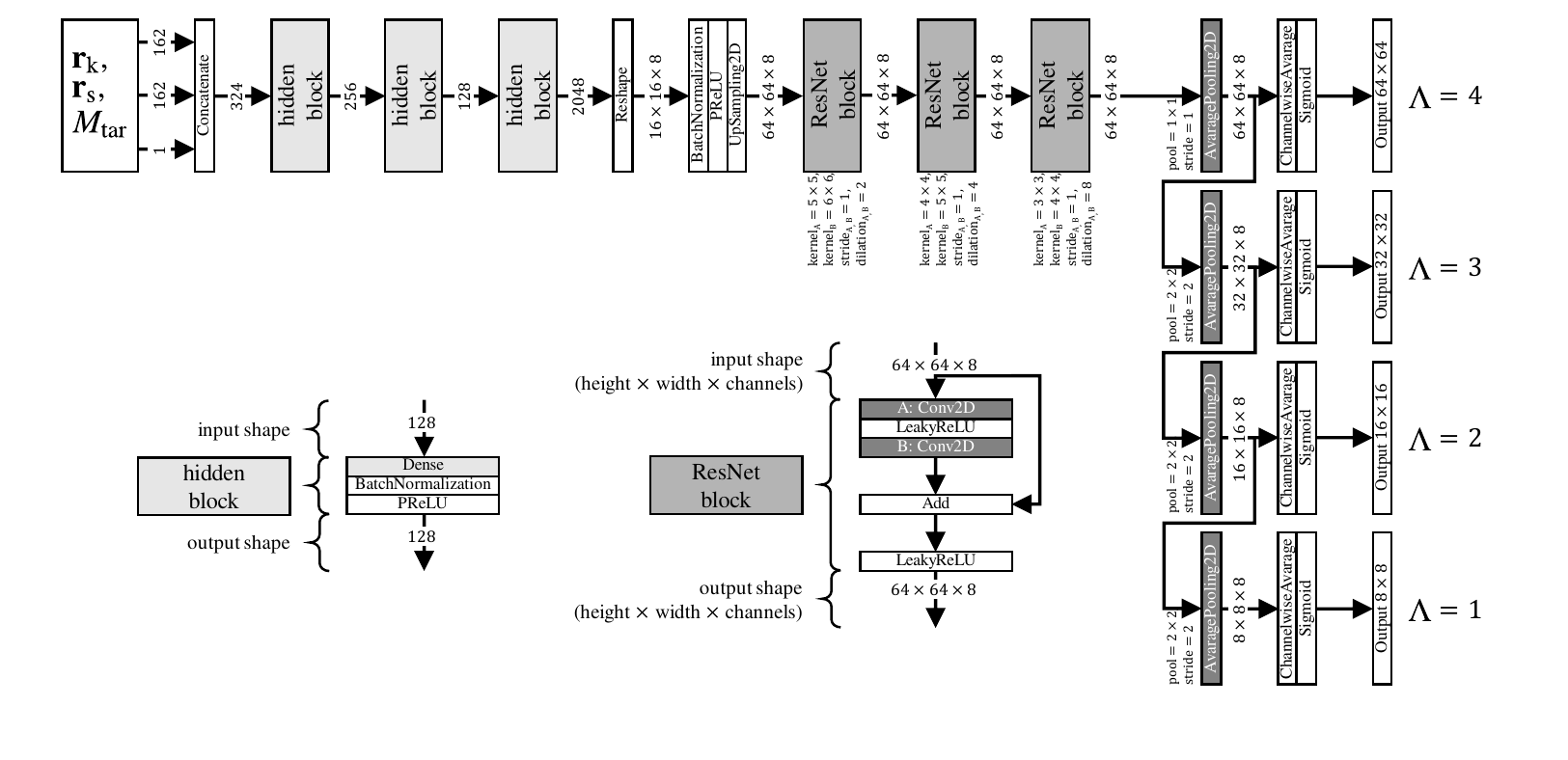}
    }
    \caption{Predictor's topology}
    \label{fig:predictor_topology}
\end{figure*}

The implementation of the presented method takes place in the programming language \hl{Python}. The framework \hl{Tensorflow} with the \hl{Keras} programming interface (API) is used, which is well suited for programming \ac{ML} algorithms in \hl{Python}. \hl{Tensorflow} is developed by Google and is an open source platform for the development of machine-learning applications \citep{Abadi2015}. In \hl{Tensorflow}, the gradients necessary for the predictor learning are calculated using \ac{AD}, which requires the use of functions available in \hl{Tensorflow} \citep{Baydin2015}. The configuration of the software and hardware used for the training is shown in Table \ref{tab:Configuration}.
\begin{table}[ht]
    \centering
    \caption{Configuration}
    \label{tab:Configuration}
    \begin{tabularx}{\linewidth}{@{}LL@{}}
        \toprule
        software or hardware & details \\
        \midrule
        GPU & Nvidia Titan RTX \\
        GDDR & 24~GB \\
        CPU & AMD Ryzen Threadripper 2950X 16-Core \\
        RAM & 128~GB \\
        Tensorflow & Version 2.5.0 \\
        Python & 3.7.3 64-bit \\
        OS & Windows 10 Version 1909 \\
        \bottomrule
    \end{tabularx}%
\end{table}

The predictor's topology, with all layers and all hyperparameters, is shown in Figure \ref{fig:predictor_topology}. The chosen hyperparameters were found to be the best after numerous tests in which the deviations of the predictions from the ones obtained by conventional \ac{TO} were evaluated. The hyperparameters are displayed by the shape (numerical expression over the arrow pointing outside the block) of the output matrix of a block or by the comment near the convolutional block. The label of the output arrow describes the dimensions of the output vector or matrix. The names of the elements in Figure \ref{fig:predictor_topology}, e.g. ``Conv2D'', correspond to the \hl{Keras} layer names.

The input data sets (top left) are processed by four fully connected (here termed ``dense'') layers, then reshaped into a three-dimensional matrix with the shape $8 \times 8 \times 64$ and passed on to two sequential convolutional layers. Subsequently, the data gets reshaped into a vector and passed through a sigmoid activation layer. As a result, the geometry at the first level is available.
The following levels build on the previous levels. So the data from the last hidden block and the data prior to the sigmoid activation of the previous level are used by for the next layer, by transforming the outputs to the same shape and adding them together. Afterwards, the data gets reshaped into a vector and, again, passed through a sigmoid activation layer. As a result, the geometry at the next level is available.

As already mentioned, the training of the predictor is based on randomly generated input data sets. All randomly chosen input data are uniformly distributed in the corresponding interval. They are generated according to the following features:
\begin{itemize}
    \item Kinematic boundary conditions $\Rk$:
    \begin{itemize}
        \item Fixed degrees of freedom along the left side in x and y direction
    \end{itemize}
    \item Static boundary condition $\Rs$:
    \begin{itemize}
            \item Position randomly chosen from all $(\dinp+1)^2=81$ nodes (except the nodes, which have a fixed degree of freedom) of level one
            \item Randomly chosen direction in the interval $\interval{0 \degree}{360 \degree}$
            \item Fixed magnitude $\RsF$
    \end{itemize}
	\item Target degree of filling $\Mtar$:
	\begin{itemize}
            \item Uniform random $\Mtar~=~\left\{0.2,0.21,\dots,0.8\right\} $.
    \end{itemize}
\end{itemize}

The algorithms \ref{pc:Learning process}, \ref{pc:Learning process part 2} and \ref{pc:Learning process part 3} show the training, the trainable parameter's update and the convergence criterion code respectively.

\begin{algorithm}[htb]
\caption{Learning process, Part 1 (training)}\label{pc:Learning process}
\begin{algorithmic}[1]
    \State $\Wp_0 \gets \text{keras initialized}$ \Comment{trainable parameters of predictor}
    \State $\vecF{X}{v}, \vecF{R}{k,v}, \vecF{R}{s,v}, M_\mathrm{tar,v} \gets \text{get validation data}$
    \State $\beta_1 = 0.9$
    \State $\beta_2 = 0.999$
    \State $\epsilon = 10^{-8}$
    \State $b \gets 0$
    \For{$\Lambda \gets 1, 4$}  \Comment{level loop} \label{pc_line: level loop}
        \State $b_{\Lambda}, \zeta_b \gets 0$
        \State $b_n \gets \vecF{B}{n}_\Lambda$
        \State $d \gets \dinp \cdot 2^{\Lambda - 1}$
        \While{$\zeta <\zeta_t$}
            \State $b, b_{\Lambda} \gets b + 1, b_{\Lambda} + 1$
            \For{$i \gets 1, b_n$} \Comment{batch loop} \label{pc_line: batch loop}
                \State $\Rk_i, \Rs_i, \Mtar_i \gets \text{random}(d)$
                \State $\vec{X}_i \gets f_{\mathrm{p}}(\Rk_i, \Rs_i, \Mtar_i, \Wp_{(b-1)}) $
                \State $c_i \gets f_{\mathrm{eva,c}}(\vec{X}_i, \Rk_i, \Rs_i)$
                \State $M_i \gets f_{\mathrm{eva,M}}(\vec{X}_i, \Mtar_i)$
                \State $\EvaC_i \gets f_{\mathrm{eva,\EvaC}}(\vec{X}_i)$
            \EndFor
            \State $J_b \gets \frac{1}{b_n} \sum_{i=1}^{b_n} f_{Q}(c_i,M_i,\EvaC_i) $
\algstore{pc_break1}
\end{algorithmic}
\end{algorithm}

The flow of data from the input $(\Rk,\Rs,\Mtar)$ of the \ac{ANN} to the output $\vec{X}$ and the objective function $J(\vec{X})$ is called forward propagation.

\begin{algorithm}[htb]
    \caption{Learning process, Part 2 (trainable parameters update)}\label{pc:Learning process part 2}
    \begin{algorithmic}[1]
    \algrestore{pc_break1}
                \State $\Gp_b \gets \nabla_{\Wp} J_b(\Wp_{b-1})$ \Comment{gradients w.r.t. objective function}
                \State $\mat{w}_b \gets \mat{w}_{b-1} - \eta(\lambda) \cdot \Gp_b$ \Comment{rough estimation, for more details see \citep{Kingma2017}}
    \algstore{pc_break2}
    \end{algorithmic}
\end{algorithm}

With computation and updating of the objective function's gradient with respect to the trainable parameters the information flows backward through the \ac{ANN}. This backward flow of information is called back-propagation \citep{Goodfellow2016}. Once the gradient is calculated, a trainable parameter's update is done using the learning rate $\eta$, which defines the length of the gradient step, and the \hl{adam} optimizer (see algorithm \ref{pc:Learning process part 2}) according to \citep{Kingma2017}.

After the trainable parameters of the predictor have been updated, a new batch can be elaborated. This process will continue until a convergence criterion is fulfilled.
In order to define a proper convergence criterion, the lowest objective function value $J_\mathrm{best}$ in the current level is tracked and compared to the current objective function value $J_b$. If the objective function value $J_b$ of one batch is not lower than $J_\mathrm{best}$ then the integer variable $\zeta_b$ (patience) is increases by one, else it resets to zero:
\begin{equation}
    \zeta_b =
    \begin{cases}
        0             & \text{if $J_b < J_\mathrm{best}$,} \\
        \zeta_{b-1} + 1 & \text{else.} \\
    \end{cases}
\end{equation}
Once the patience exceeds a predefined value $\zeta_\mathrm{max}$, termed maximal patience (see table \ref{tab:Parameter}), the level $\Lambda$ increases or, if the maximum level was reached, the training stops (see algorithm \ref{pc:Learning process} line \ref{pc_line: level loop}).

\begin{algorithm}[ht]
    \caption{Learning process, Part 3 (convergence criterion)}\label{pc:Learning process part 3}
    \begin{algorithmic}[1]
    \algrestore{pc_break2}
                \If{$b_{\Lambda} = 1$}
                    \State $J_\mathrm{best} \gets J_b$
                \EndIf
                \If{$J_b \geq J_\mathrm{best}$}
                    \State $\zeta_b \gets \zeta_{b-1} + 1$
                \Else
                    \State $\zeta_b \gets 0$
                \EndIf
            \EndWhile
        \EndFor
    \end{algorithmic}
\end{algorithm}

The parameters in table \ref{tab:Parameter} were used for the training.
\begin{table}[htb]
    \small
    \caption{Parameter}
    \label{tab:Parameter}
    \begin{tabularx}{\linewidth}{@{}llX@{}}
        \toprule
        Symbol & Value & Description \\
        \midrule
        $\alpha$ & 2     & $f_\mathrm{Q}$ coefficient for $c$ \\
        $\beta$ & 5      & $f_\mathrm{Q}$ coefficient for $M$ \\
        $\gamma$ & 1     & $f_\mathrm{Q}$ coefficient for $\EvaC$ \\
        $\delta$ & 1     & $f_\mathrm{Q}$ coefficient for $P$ \\
        $\zeta_\mathrm{max}$ & 1000  & maximal patience \\
        $\eta(\lambda)$ & $\frac{1}{4}^{(\lambda - 1)}\frac{1}{100}$ & learning rate for different levels $\Lambda$\\
        $\nu$ & 0.3   & Poisson's ratio used for conventional \ac{TO} \\
        $\vec{b}_n^\T$ & \multicolumn{1}{p{5.355em}}{\batchSizesperLevel{}} & batch size of the geometry for different levels $\Lambda$ \\
        $\dinp$ & 8     & size of geometry in the first level along one side \\
        $E$   & 195000 $\mathrm{\frac{N}{mm^2}}$ & Young's modulus \\
        $\EvaC_{\mathrm{k}}$ & 3     & parameter for $\EvaC$ \\
        $p$   & 3     & power penalization factor used for conventional \ac{TO} \\
        $r_\mathrm{min}$ & 3     & filter size \\
        $\RsF$ & 100 $\mathrm{N}$ & magnitude of the forces of the static boundary condition \\
        $\Xmin$ & 0.001 & lower limit value for the entries of $\vec{X}$ \\
        \bottomrule
    \end{tabularx}%
\end{table}

        \subsection{Results}
\label{sec:Results}

\begin{figure}[ht]
    \centering
    \includegraphics[width=78mm]{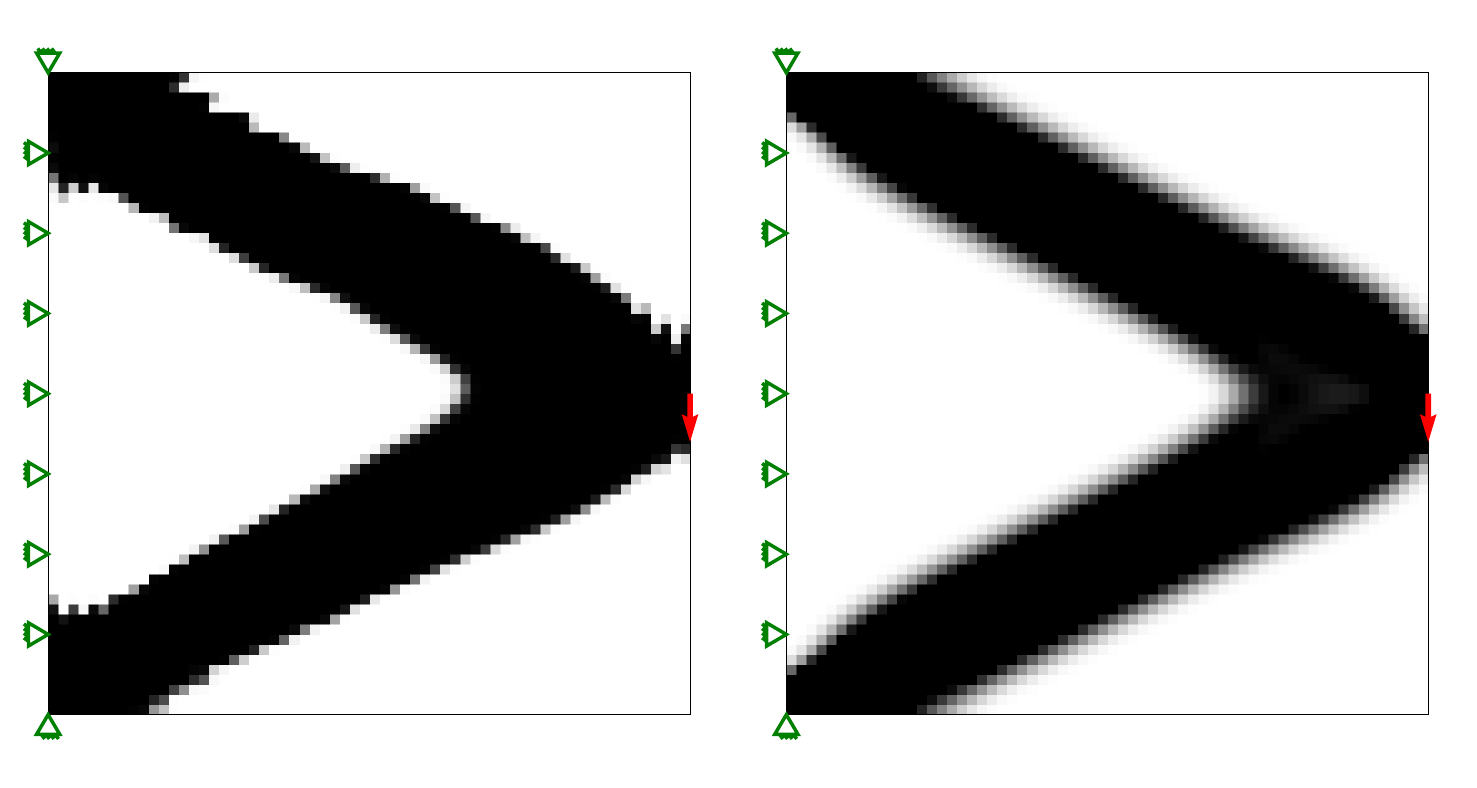}
    \caption{Sample geometry (Left: \ac{ANN}-based \ac{TO} by using the \ac{PEN}-Method; Right: \acl{top88} \citep{Andreassen2011})}
    \label{fig:Sample geometry}
\end{figure}

The training of the predictor lasted \trainingduration{}, which can be subdivided according to the individual levels as follows: \trainingsubdivision{}. The \ac{ANN} based \ac{TO} geometries are similar to the results obtained by  \ac{top88} according to \citep{Andreassen2011} for the same input datasets.
For the conventional \ac{SIMP}-\ac{TO} the density filter method and the parameters mentioned in table \ref{tab:Parameter} were used.
An example prediction is shown in Figure \ref{fig:Sample geometry}.

\begin{figure}[ht]
    \centering
    \includegraphics[width=78mm]{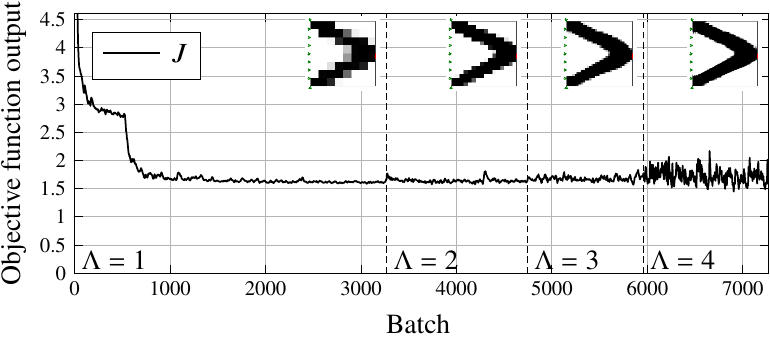}
    \caption{Learning progress: objective function}
    \label{fig:Learning progress: objective function}
\end{figure}

\begin{figure}[ht]
    \centering
    \includegraphics[width=78mm]{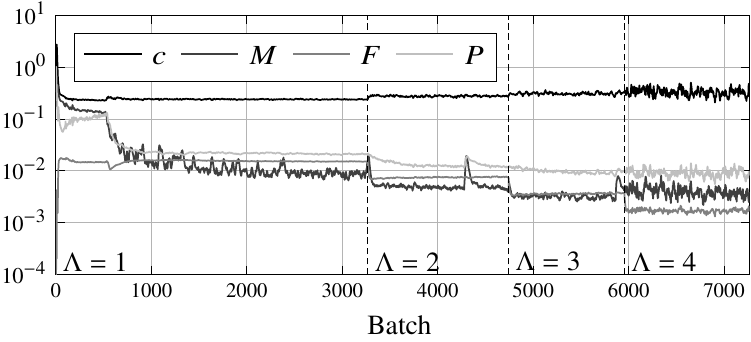}
    \caption{Learning progress: evaluator losses}
    \label{fig:Learning progress: evaluators}
\end{figure}

The training history shows the progression of the objective function (see Figure \ref{fig:Learning progress: objective function}) and of the individual evaluator losses over the number of batches (see Figure \ref{fig:Learning progress: evaluators}).
The smaller batch size at higher levels produces more oscillation of the curve and therefore makes it difficult to identify a trend.
For this reason, the curves shown in the figures are filtered using the exponential moving average and a smoothing factor of \exponentialsmoothingfactor{} \citep{Nicolas2017}.
This filtering does not affect the original objective function and serves only for visual purposes.

The dashed vertical lines (labeled with the value of $\Lambda$) in the Figure \ref{fig:Learning progress: objective function} and \ref{fig:Learning progress: evaluators} show the change of level.
It can be seen that after each increase of $\Lambda$, the value of the objective function increases.  
This can be explained by adding more weights that are randomly distributed and still untrained. 

The results were validated using $n=100$ randomly generated input data sets, called validation data, that were not part of the training data sets, and the corresponding optimized geometries which were conventionally calculated by the \ac{top88} available in \citep{TOPOPTgroup2020}.

\begin{figure}
    \centering
    \subfloat[computing time]{{
        \includegraphics[scale=1,valign=t]{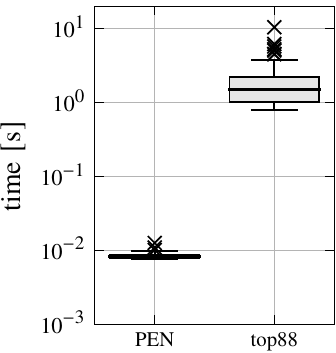}%
        \vphantom{\includegraphics[scale=1,valign=t]{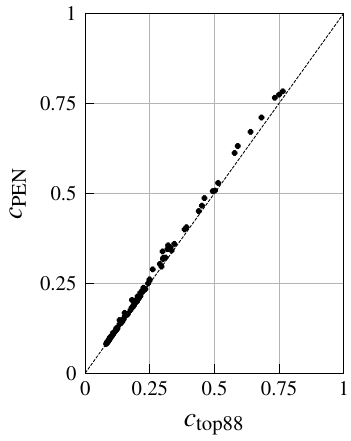}}%
    }}
    \subfloat[compliance]{{
        \includegraphics[scale=1,valign=t]{fig_14b.pdf}%
    }}
    \caption{Computing time and compliance comparison}
    \label{fig:Result comparison}
\end{figure}

The results of the comparison (\ac{PEN} and \ac{top88}) of the 100 validation data sets are summarized in the plots in Figure \ref{fig:Result comparison}.

On average, the \ac{ANN}-based \ac{TO} can deliver almost the same result as the conventional method in about \timePENperGeometry{}, while the conventional topology optimizer according to Andreassen \citep{Andreassen2011} requires on average \timeCONVperGeometry{} (and is hence roughly \timeCONVslowerPEN{} times slower), see Figure \ref{fig:Result comparison} a). It can also be seen that the majority of geometries generated by \ac{PEN} have a compliance that is close to the geometries generated by \ac{top88}, see Figure \ref{fig:Result comparison} b).

\begin{figure*}[bt]
    \centering
    \includegraphics[width=\textwidth]{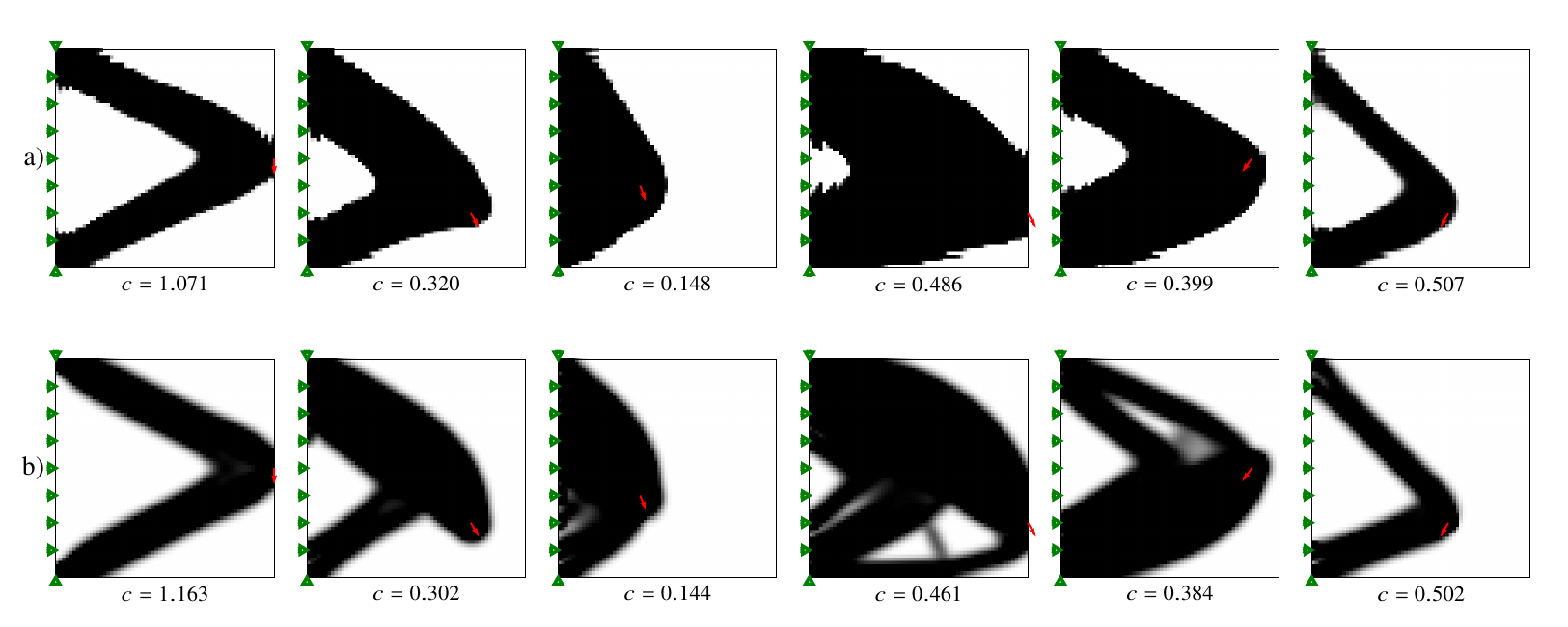}
    \caption{Additional sample geometries a) Deep Learning Topology Optimization
        b) validation data
    }
    \label{fig:Further sample geometries}
\end{figure*}

In addition to the comparison of compliance and computing time, there are indicators that allow the comparison of geometries generated by \ac{ANN} based \ac{TO} with geometries generated by conventional \ac{TO}.

The accuracy $\kappa$ represents the conformity of the predicted geometries with the validation data set geometries. To determine the deviation of a single validation geometry $\vecF{X}{v}$ from the predicted geometry $\vecF{X}{p}$, obtained from same input datasets, the functions
\begin{equation}
    mae(\vecF{X}{p}, \vecF{X}{v}) =
    \frac{1}{n} \sum_{i=1}^n \left\lvert \vecEF{X}{p}_i - \vecEF{X}{v}_i \right\rvert
    \label{eq:mae}
\end{equation}
and
\begin{equation}
    mse(\vecF{X}{p}, \vecF{X}{v}) =
    \frac{1}{n} \sum_{i=1}^n (\vecEF{X}{p}_i - \vecEF{X}{v}_i)^2
    \label{eq:mse}
\end{equation}
are used, where $n$ represents the number of validation data sets and in this case has the value 100. The smaller the \ac{mse} and \ac{mae} values are, the better the similarity of the results. The function $\kappa_{0.01}$ indicates the percentage of elements with density differences of less than or equal to $0.01$ with respect to the density values of the validation geometries generated by the conventional TO. For the calculation of $\kappa_{0.01}$ the Heaviside step function $\Theta$ is used:
\begin{equation}
    \kappa_{0.01}(\vecF{X}{p}, \vecF{X}{v}) =
    \frac{1}{n} \sum_{i=1}^n \Theta\left(1- \frac{\left\lvert \vecEF{X}{p}_i - \vecEF{X}{v}_i \right\rvert}{0.01} \right)
    .
    \label{eq:nu_0.01}
\end{equation}

With the help of the function
\begin{equation}
    \begin{aligned}
    \kappa(\vecF{X}{p}, \vecF{X}{v}) =
    \frac{1}{n} \sum_{j=1}^{n} \frac{1}{3} & \left[
        \left( 1 - mse(\vecEF{X}{p}_i, \vecEF{X}{v}_i) \right) \right. \\
        & + \left. \left(1 - mae(\vecEF{X}{p}_i, \vecEF{X}{v}_i) \right) \right. \\
        & + \left. \kappa_{0.01}(\vecEF{X}{p}_i, \vecEF{X}{v}_i) \right]
    \end{aligned}
    \label{eq:accur}
\end{equation}
the accuracy and thus the validity of the predictor can be determined and predictors with different hyperparameters can be compared. The function $\kappa$ is needed because a single indicator is not sufficient to determine the accuracy of the predictor. So it is possible that the accuracy $\kappa_{0.01}$ is low and the errors \ac{mae} and \ac{mse} are also small at the same time. Since these error indicators concentrate on different kinds of differences, the average $\kappa$ of those is a more meaningful indicator.

The examples in Figure \ref{fig:Further sample geometries} show that the predictor can deliver geometries that are similar to conventional method as well as some weaknesses. 
For example, in some cases the geometries are noisy and contain undesirable elements, which do not contribute to the stiffness (see Figure \ref{fig:Further sample geometries}, column two or four). This may be improved by an appropriate choice of hyperparameters of the predictor and by adapting the quality function. Also included in this figure is a row of conventionally topology optimized geometries using different parameters. For all sample geometries in Figure \ref{fig:Further sample geometries} the compliance is reported under the geometry diagram. For the \ac{ANN} \ac{TO} generated geometries the evaluator losses are summarized in table \ref{tab:Summary of evaluator outputs}.

\begin{table}[ht]
    \small
    \caption{Summary of evaluator losses (same examples as in Figure \ref{fig:Further sample geometries})}
    \label{tab:Summary of evaluator outputs}
    \begin{tabularx}{\linewidth}{@{}l|XXXXXX@{}}
        \toprule
        Example    & 1     & 2     & 3     & 4     & 5     & 6 \\
        \midrule
        $c$   & 1.071 & 0.320 & 0.148 & 0.486 & 0.399 & 0.507 \\
        $M$   & 0.003 & 0.003 & 0.001 & 0.012 & 0.003 & 0.009 \\
        $F$   & 0.003 & 0.003 & 0.001 & 0.003 & 0.003 & 0.002 \\
        $P$   & 0.013 & 0.007 & 0.006 & 0.005 & 0.007 & 0.017 \\
        \bottomrule
    \end{tabularx}%
\end{table}

The table \ref{tab:Summary of indicators} shows all kinds of indicators that represent the conformity of \ac{PEN}-method generated geometries with conventionally generated geometries, averaged over all 100 examples. \acs{SD} is the \acl{SD}.

\begin{table}[ht]
    \small
    \caption{Summary of indicators}
    \label{tab:Summary of indicators}
    \begin{tabularx}{\linewidth}{@{}l|XXXX@{}}
        \toprule
                        & $mse$  & $mae$  & $\kappa_{0.01}$ & $\kappa$ \\
        \midrule
        Mean            & \msemean{}      & \maemean{}      & \kappaOOImean{}          & \kappamean{} \\
        \ac{SD}         & \mseSD{}        & \maeSD{}        & \kappaOOISD{}           & \kappaSD{} \\
        \bottomrule
    \end{tabularx}%
\end{table}

From the data in the table \ref{tab:Summary of indicators} it can be seen that in the examined cases \kappaOOImean{} of the elements of the geometries obtained with the \ac{PEN} method have density differences of less than 1\% as compared to the conventionally optimized geometries.

        \subsection{Computing time comparison}
\label{sec:Computing time comparison}

As mentioned in section \ref{sec:Results} the \ac{PEN} method is by orders of magnitude faster than \ac{top88}. However, the predictor profits from a computationally intensive training. So it is interesting to attempt a comparison which takes into account the training time.

The \ac{PEN} computing time for a single geometry $t_\mathrm{PEN}$, including its share of training time, obviously depends from the number of geometries predicted $e_{\mathrm{p}}$ on the basis of one single training process:
\begin{equation}
    t_{\mathrm{PEN}} = \frac{T_{\mathrm{p}}}{e_{\mathrm{p}}}+t_{\mathrm{P}}
\label{eq:t_{PEN}}
\end{equation}
where $t_{\mathrm{P}}$ is the computing time per single geometry and $T_{\mathrm{p}}$ is the training time. The \ac{BEP} is given by the number of predictions $e_{\mathrm{BEP}}$ for which both methods require the same time (including training time contribution). To calculate the \ac{BEP} $t_{\mathrm{PEN}}$ is set equal to $t_{\mathrm{TO}}$, which is the computing time to optimize a single geometry using the conventional method. It results:
\begin{equation}
    e_{\mathrm{BEP}} = \frac{T_{\mathrm{p}}}{t_{\mathrm{TO}}-t_{\mathrm{p}}}
    \label{eq:BEP}.
\end{equation}

The table \ref{tab:Time comparison of conventional and PEN based methods} shows the computing times of the different methods as well as the \ac{BEP}.

\begin{table}[htb]
    \small
    \begin{tabularx}{\linewidth}{@{}XXX@{}}
        \toprule
        $t_\mathrm{p}$ & $t_\mathrm{TO}$ & $e_{\mathrm{BEP}}$ \\
        \midrule
        \timePENperGeometry & \timeCONVperGeometry & \timeBEP \\
        \bottomrule
    \end{tabularx}%
    \caption{Time comparison of conventional and PEN based methods}
    \label{tab:Time comparison of conventional and PEN based methods}
\end{table}

When evaluating the results of this comparison, the following points should be considered:
\begin{itemize}
    \item Due to the fact that $t_{\mathrm{p}} \ll t_{\mathrm{TO}}$ the \ac{BEP}, for a given reference method, essentially depends on the training time.
    \item The training time in turn depends on the convergency condition. Within the framework of this project an extensive study about the proper of choice of convergency criterion could not be made. The present choice allowed for good results. It can be expected that the training time could be reduced after a targeted study in this sense.
    \item Of course the training time also depends on the hardware used for training. By using a high performance hardware the training time, and so the \ac{BEP} can be strongly reduced without effecting the versatility of the method in everyday use.
    \item This comparison does not include a study of the effect of the problem size (number of design variables).
\end{itemize}

        \subsection{Online}
\label{sec:Online}
Due to the ability to quickly get the optimized geometry by the predictor, the \ac{ANN}-based \ac{TO} can be executed online in the browser. Under the address: \url{https://www.tu-chemnitz.de/mb/mp/forschung/ai-design/TODL/} it is possible to perform investigations with different degrees of filling as well as static boundary conditions.

    \section{Conclusion}
\label{sec:Conclusion}

In this paper, a method was presented that makes it possible to realize a topology optimizer using deep learning. The \ac{ANN} in charge of generating topology-optimized geometries does not need any pre-optimized data sets for the training. The generated geometries are in most cases very similar to the results of conventional topology optimization according to Sigmund or Andreassen.

This topology optimizer is much faster, due to the fact that the computing-intensive part is shifted into the training. After the training, the \acl{ANN} based topology optimizer is able to deliver geometries which are nearly identical to the ones generated by conventional topology optimizers. This is achieved by using a new approach, the \acf{PEN} approach. \ac{PEN} consists of a trainable predictor, which is in charge of generating geometries, and evaluators, which have the purpose of evaluating the output of the predictor during the training.

The method was tested up to an output resolution of $64 \times 64$. The optimization of the computational efficiency of the training phase was not the first priority of this project since the training is performed just once and therefore affects the performance of the method only in a limited fashion.
A critical step is the calculation of the displacements in the compliance evaluator. The use of faster algorithms (e.g. sparse solvers) could remove the mentioned limitations.
One improving option could consist in implementing the compliance evaluator as an \ac{ANN} itself and thus making it faster and more memory-efficient. This would make possible to cope with finer resolutions or to learn with much larger batch sizes and thus with more training data in the same time.

The results of the \ac{PEN} method are comparable to the ones of the conventional method. However, the \ac{PEN} method could prove superior in handling applications and optimization problems of higher complexity, such as stress limitations, compliant mechanisms and many more. This expectation is related to the fact that no optimized data are needed. All methods which process pre-optimized data suffer from the difficulties encountered by conventional optimization while managing the above-mentioned problems. Because the \ac{PEN} method works without optimized data, it can also be applied to problems that have no optimal solutions or solutions that are hard to calculate, like the fully stressed truss optimization.

Up to now, variable kinematic boundary conditions were not tested. This will be done in future research, together with resolution improvement, application to three-dimensional design domains and consideration of nonlinearities and restrictions.

    \section{Replication of Results}
\label{sec:Replication of Results}

The \hl{Keras}-PEN network can be tested at \url{https://www.tu-chemnitz.de/mb/mp/forschung/ai-design/TODL/}. The download of the model, as well as the download of the validation data is available at \url{http://dx.doi.org/10.17632/459f33wxf6.2} \cite{Halle2020a}.

    \printbibliography
\end{document}